\documentclass{article}

\usepackage[dblblindworkshop, final]{neurips_2026}

\usepackage[utf8]{inputenc} 
\usepackage[T1]{fontenc}    
\usepackage{hyperref}       
\usepackage{url}            
\usepackage{booktabs}       
\usepackage{graphicx}       
\usepackage{amsfonts}       
\usepackage{nicefrac}       
\usepackage{microtype}      
\usepackage{xcolor}         
\usepackage{pifont}
\usepackage{tabularx}
\usepackage{subcaption}

\PassOptionsToPackage{numbers, compress}{natbib}

\definecolor{darkblue}{HTML}{1B76D2}

\workshoptitle{Representations for the Physical Sciences}
\title{Do Tabular Foundation Models Know Physics? Contamination, Units, and the Deterministic Limit}

\author{%
  Wassim Tenachi \\
  D\'epartement de physique, Universit\'e de Montr\'eal \\
  Mila -- Quebec Artificial Intelligence Institute \\
  \texttt{wassim.tenachi@mila.quebec} \\
  \AND
  Yashar Hezaveh \\
  D\'epartement de physique, Universit\'e de Montr\'eal \\
  Mila -- Quebec Artificial Intelligence Institute \\
  \texttt{yashar.hezaveh@mila.quebec} \\
  \AND
  Laurence Perreault Levasseur \\
  D\'epartement de physique, Universit\'e de Montr\'eal \\
  Mila -- Quebec Artificial Intelligence Institute \\
  \texttt{levassel@mila.quebec} \\
  \AND
  Pierre-Luc Bacon \\
  Department of Computer Science and Operations Research, Universit\'e de Montr\'eal \\
  Mila -- Quebec Artificial Intelligence Institute \\
  \texttt{pierre-luc.bacon@mila.quebec} \\
}

\begin{document}

\maketitle

\begin{abstract}
Tabular foundation models (TFMs) learn to fill in tables the way language models fill in text --- and tables are arguably the format in which most physical measurement arrives. Did they learn any physics in the process? They are Bayesian by construction, so the question is what their prior contains. We probe it directly, evaluating four of them (\texttt{TabPFN-3}, \texttt{TabICLv2}, \texttt{TabDPT} and \texttt{Real-TabPFN-2.5}) against six baselines on datasets sampled from 316 physical equations, in and out of domain. TFMs dominate, out of the box and after tuning. But they struggle to represent noiseless deterministic mechanisms in their uncertainty, and show no sign of exploiting physical units: they interpolate physics without yet being able to act as physical models.
\end{abstract}



\begin{figure}[h!]
  \centering
  \includegraphics[width=1.\hsize]{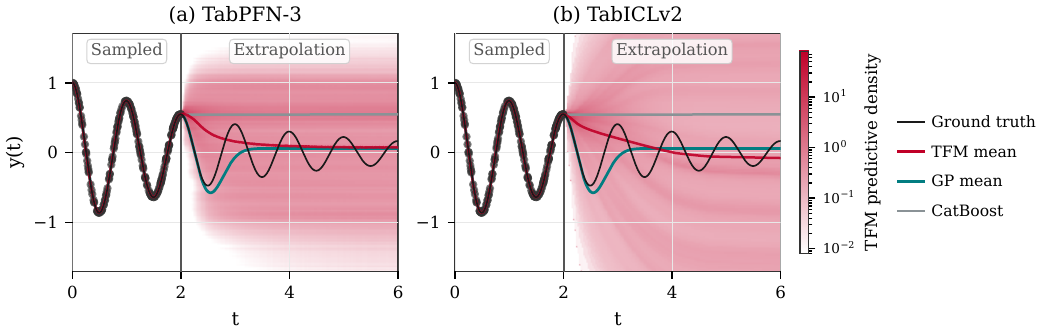}
  \caption{\textbf{TFMs interpolate physics but do not learn enough of it to extrapolate.} A noiseless damped harmonic oscillator queried beyond the sampled range, against baselines. The oscillation is recovered where there is data and lost immediately outside it, though the TFM means still settle near the equilibrium the oscillator decays to, rather than holding the last observed value. Details and other models in App.~\ref{app:dho}.}
  \label{fig:dho}
\end{figure}


\section{Introduction}
Nature, in Galileo's phrase, is written in the language of
mathematics~\citep{Galilei1623}. We do not read it directly, however: we observe
it as data, and in the physical sciences that data is overwhelmingly tabular~\citep{VizieR}. Large language models have made striking progress on
the mathematical language itself~\citep{OpenAIconjecture}, and vision models on
scientific imaging~\citep{AION}. Tables, arguably the format in which most
physical measurement actually arrives, have received far less attention.
Skimming oceans of text produced powerful representations of the physical
world~\citep{Gemini}. Would skimming oceans of tables do the same?

\textbf{Tabular foundation models (TFMs)} learn to fill in tables much as
language models learn to fill in text: entries are masked and the model is
trained to recover them from context~\citep{TabPFN_ICLR}. They are invariant to
row and column permutations, and most emit a full predictive distribution per
cell. A new task is therefore solved in a single forward pass --- in-context
learning (ICL), with no gradient step --- so fitting is free, amortized by
pretraining, while querying costs grow with context size. Most are pretrained
exclusively on synthetic structural causal models (SCMs)\footnote{Random graphs in which each node is generated from its parents plus independent
noise.}~\citep{TabPFN_ICLR, TabICL}; others on corpora of real
tables~\citep{TabDPT, RealTabPFN}.

\textbf{Related work.} TFMs are now applied across scientific fields~\citep{TabPFN3}, including astrophysics~\citep{QuasarTabFM},
though almost exclusively in interpolation. Benchmarking efforts such as
\texttt{TabArena}~\citep{TabArena} and \texttt{BeyondArena}~\citep{BeyondArena}
have concentrated on stochastic processes, and have pushed the field toward
extrapolation, where TFMs remain weak --- strikingly so on physical signals
(Fig.~\ref{fig:dho}). But how should a model extrapolate an arbitrary real-world table without having
learned something of the physics that generated it?

\textbf{Two obstacles.} Physics-equation data sits close enough to the ecosystem
these models are trained from\footnote{Feynman-equation datasets make up
${\sim}118$ of the 1053 datasets surveyed as candidates for
\texttt{TabArena}, mostly via \texttt{PMLB}~\citep{PMLB}; the final 51-dataset
suite excludes them.} to warrant a contamination audit before interpreting any
result (\S\ref{subsec:contamination}). Yet the physics regime is itself absent
from the literature: \texttt{BeyondArena} drops deterministic-function data as
artificial, and no current TFM has been pretrained on a physical law with units
or a continuous target (App.~\ref{app:tfm_training_sets}).

\textbf{Which prior?} TFMs are Bayesian by construction: pretraining on samples
from a prior under a proper scoring rule makes the forward pass an
approximation to the posterior predictive~\citep{TabPFN_Nature}. They return
posteriors, not point estimates. The question is what prior those posteriors are
built on, and whether it has anything to do with physics --- and data sampled from known physical laws is the cleanest way to ask.

\section{Protocol}
\label{sec:protocol}

\textbf{Data.} We evaluate on three sets of data sampled from physical
equations: Feynman (120 equations)~\citep{AIFeynman}, standardized in
\texttt{SRBench}~\citep{SRBench}, and \texttt{LSR-Transform} (111) and
\texttt{LSR-Synth} (85)~\citep{LSR}, which rewrite Feynman into unusual forms and
compose novel terms respectively. Every table is regenerated from its equation
rather than loaded, so no published table is reused and $n_\mathrm{samples}$,
noise and sampling domain stay free.

\textbf{Regimes.} Training points are drawn from each equation's sampling box, test points come either from the same box ($k{=}1$, in-domain) or from a shell outside it at twice its half-width ($k{=}2$, extrapolation). Targets carry multiplicative noise
$y'=y(1+\varepsilon)$, $\varepsilon\sim\mathcal{N}(0,\sigma^2)$, emulating the
relative error of physical instruments. We sweep
$n_\mathrm{samples}\in\{50,200,2000\}$, $\sigma\in\{0,0.01,0.1\}$, $k\in\{1,2\}$
and 3 seeds, with the test set fixed at 2000 points throughout. The detailed protocol is given in
App.~\ref{app:protocol}.

\textbf{Baselines and evaluation.} Following
\texttt{BeyondArena}~\citep{BeyondArena}, we evaluate the four open-weight TFMs (Table~\ref{tab:tfms}) against six trained baselines: \texttt{RealMLP},
\texttt{TabM}, a vanilla \texttt{MLP},
\texttt{CatBoost}, a Gaussian process (\texttt{GP}) and \texttt{Ridge} (App.~\ref{app:baselines}). We score with
$\mathrm{NMSE}=\mathbb{E}[(y-\hat y)^2]/\mathbb{E}[(y-\bar y)^2]$; since raw
error spans orders of magnitude across equations, we rank models within each
(task, regime) cell and report mean rank per stratum.

\textbf{Fairness.} Comparing a forward pass to a training loop requires fixing a
budget, so we report both. (i) Under \emph{default}, every method uses library
defaults. (ii) Under \emph{tuned}, the trained baselines get 25 random-search
configurations, optionally combined by post-hoc greedy ensemble
selection\footnote{Caruana-style~\citep{Caruana2004} forward selection with
replacement over a model's own 25 configurations, scored on validation; no
additional fits are needed.}; TFMs have no hyperparameters, and their analogue
is the ensemble size $n_{\mathrm{estimators}}$, which averages the forward pass
over input permutations. We report both families at their cheapest and strongest
settings.

\begin{table}[t]
  \centering
  \small
  \caption{\textbf{Evaluated tabular foundation models.} \emph{SCMs}: pretrained on synthetic tables drawn from structural causal models. \emph{Real data}: pretrained on real-world tables. \emph{Reg.}: those real tables include regression targets, as opposed to classification only. Only \texttt{TabDPT} has been pretrained on real-world regression data.}
  \resizebox{0.8\hsize}{!}{%
  \begin{tabular}{llcccl}
    \toprule
    Model & Output & SCMs & Real data & Reg. & Ref. \\
    \midrule
    \texttt{TabPFN-3}        & Quantiles & \checkmark & $\times$   & ---        & \citep{TabPFN3} \\
    \texttt{TabICLv2}        & Quantiles & \checkmark & $\times$   & ---        & \citep{TabICL} \\
    \texttt{Real-TabPFN-2.5} & Quantiles & \checkmark & \checkmark & $\times$   & \citep{RealTabPFN} \\
    \texttt{TabDPT}          & Point     & $\times$   & \checkmark & \checkmark & \citep{TabDPT} \\
    \bottomrule
  \end{tabular}%
  }
  \label{tab:tfms}
\end{table}

\section{Results \& Analysis}
\label{sec:analysis}

\begin{figure}[h!]
  \centering
  \includegraphics[width=1.\hsize]{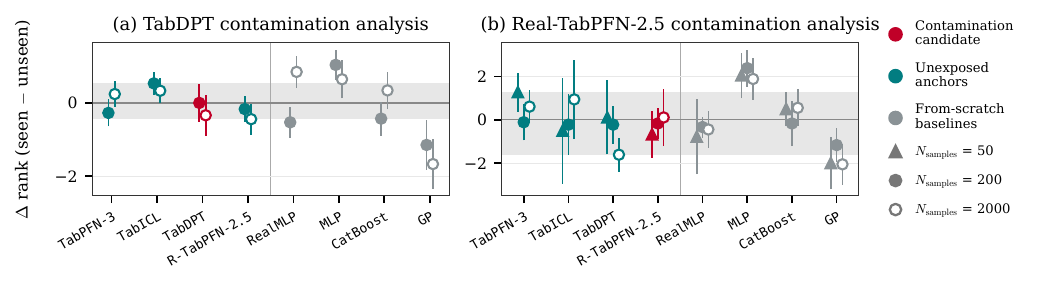}
  \caption{\textbf{No detectable contamination from real-data pretraining.}
  Difference in mean rank between datasets seen during pretraining and comparable unseen datasets, for \textbf{(a)} TabDPT and \textbf{(b)} Real-TabPFN-2.5. Negative values indicate better performance on the seen trained data. The shaded band spans the unexposed TFM anchors; both candidates fall inside it at every context size.}
  \label{fig:contamination}
\end{figure}

\paragraph{No detectable contamination.}
\label{subsec:contamination}
Before asking whether TFMs know physics, we must rule out that they have simply
seen our evaluation data. We audited the pretraining corpora of the two
real-data models, both of which enumerate them: \texttt{TabDPT}'s 122 datasets
and \texttt{Real-TabPFN-2.5}'s 43. Neither contains a Feynman equation, nor any
regression target generated by an analytic function. Physics appears only as
detector classification, never as
law\footnote{\href{https://www.openml.org/search?type=data&id=40679}{\texttt{MAGIC}},
\href{https://www.openml.org/search?type=data&id=23512}{\texttt{HIGGS}} and
\href{https://www.openml.org/search?type=data&id=41150}{\texttt{MiniBooNE}} on
OpenML.} (full corpora in App.~\ref{app:tfm_training_sets}).

A list of names is not a list of generating processes, however: both corpora
were filtered against benchmark suites by dataset name and content hash, and a
table resampled from the same equation under a different name would pass both
checks. We therefore measure exposure rather than assert it, comparing each
model on datasets drawn from its own corpus against comparable datasets outside
it, and reading the gap relative to models that saw
neither\footnote{For \texttt{TabDPT}, the 28 regression datasets in its corpus
against 37 matched datasets from \texttt{TabArena} and OpenML-CTR23. For
\texttt{Real-TabPFN-2.5}, the single function-generated table it has seen
(\texttt{fried}, and only in binarized form) against \texttt{friedman2}
\citep{PMLB}, an unexposed member of the same family --- one dataset, hence the
noisier panel.} (Fig.~\ref{fig:contamination}).

Both come back null: each candidate sits inside the unexposed band at every
context size, and what little movement exists runs counter to the mechanistic
prediction, growing with context size where prior exposure should matter most
when context is thin. Most tellingly, models with no pretraining at all show
\emph{larger} seen-versus-unseen gaps than the exposed models do, so this
statistic is dominated by which datasets fall in each pool rather than by
exposure to them. That spread also bounds what we could have detected: any
contamination effect below roughly one rank position would be invisible at this
pool size.


\begin{figure}[h!]
  \centering
  \includegraphics[width=1.\hsize]{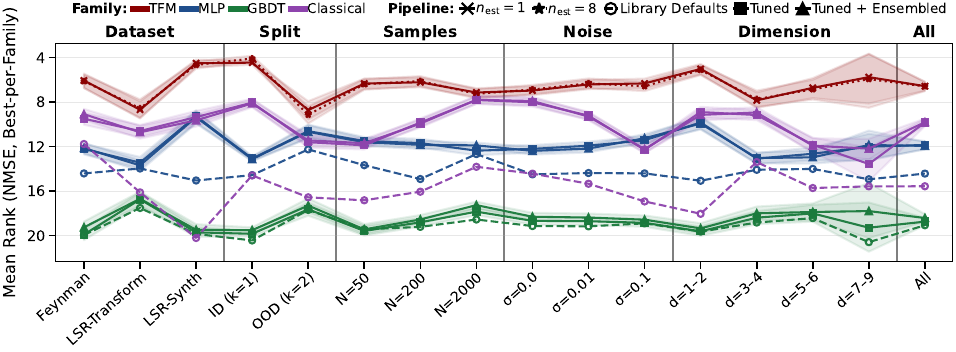}
\caption{\textbf{The TFM lead on physics data survives tuning and ensembling.}
Mean rank within each stratum (lower is better). Details and per-model results in App.~\ref{app:phyarena_full} –– \texttt{TabPFN-3} leads throughout.}
  \label{fig:phyarena}
\end{figure}

\paragraph{Tuning does not close the gap.}
Under default configurations our ordering reproduces \texttt{BeyondArena}'s:
TFMs lead the trained baselines across every stratum
(Fig.~\ref{fig:phyarena}). Their reported reversal arises from tuning, which we
therefore evaluate directly. It does not occur here. Per model
(Fig.~\ref{fig:phyarena_full}), \texttt{TabPFN-3} still leads under all three
pipelines in every stratum, the sole exception being the Gaussian process at the
largest context size, and only against the TFM's cheapest setting. It also holds
where both conditions that favor the trained models apply at once --- out of
domain at the largest context size --- though the margin narrows
there\footnote{Mean rank over 314 tasks: $8.47$ for \texttt{TabPFN-3} against
$9.85$ for the tuned and ensembled GP, the best baseline here.}. The clearest
breakdown in ordering comes not from tuning but from extrapolation: out of
domain the field compresses and the ranking reshuffles more than anywhere else
(Fig.~\ref{fig:phyarena_full}), models converging on a common failure rather
than degrading in place.

\paragraph{Can the prior represent a noiseless mechanism?} \label{subsec:physics}
TFMs are pretrained on synthetic data in which every variable carries
independent noise, how do they behave on noiseless data? On data drawn from a
deterministic law $f$, aleatoric noise is zero, so in interpolation the correct
posterior predictive approaches a delta function as context grows. At finite
context its width is not zero, since several functions still fit the observed
points, but it should shrink as the predictions improve. A well-calibrated
model's 80\% interval should therefore keep covering 80\% of test points as
$n_\mathrm{samples}$ grows. 

\begin{figure}[ht]
  \centering
  \begin{subfigure}{0.58\hsize}
    \includegraphics[width=\hsize]{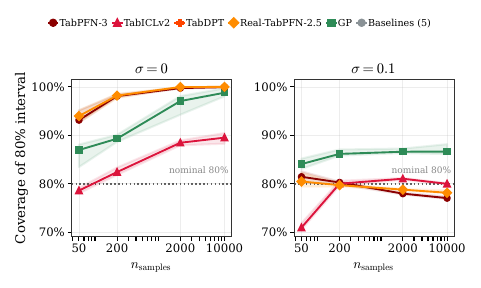}
    \caption{\textbf{TFMs are calibrated with noise but underconfident without it.} Fraction of in-domain test points inside each model's 80\% predictive interval (median over tasks), a calibrated model stays on the dotted line. At $\sigma{=}0.1$ the TFMs sit near it beyond the smallest context, at $\sigma{=}0$ their coverage climbs toward 100\% as context grows, most strongly for the two \texttt{TabPFN} models.}
    \label{fig:detlim}
  \end{subfigure}
  \hfill
  \begin{subfigure}{0.40\hsize}
    \includegraphics[width=\hsize]{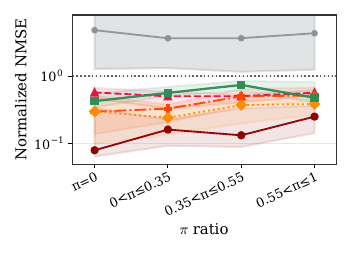}
    \caption{\textbf{Dimensional structure leaves little trace on error.} Median NMSE relative to the typical model on each task (below 1 is better), binned by the fraction of input variables left as dimensionless groups, $n_\pi/n_\mathrm{vars}$; the $n_\pi{=}0$ tasks are all pure power laws. Other models vary less than \texttt{TabPFN-3}, whose lead is largest on the most reducible tasks.}
    \label{fig:units}
  \end{subfigure}
  \label{fig:prior}
\end{figure}

In the noisy regime ($\sigma{=}0.1$), the TFMs stay
close to 80\% (Fig.~\ref{fig:detlim}-right). In the noiseless regime
($\sigma{=}0$), their coverage instead climbs toward 100\%
(Fig.~\ref{fig:detlim}-left): the intervals shrink more slowly than the error,
so the models become underconfident, as if a residual noise floor kept their
predictive distributions too wide. The effect is strongest for the two
\texttt{TabPFN} models, near 100\% from $n_\mathrm{samples}{=}200$, and mild
for \texttt{TabICLv2}. The GP shows it later, as expected: its fitted noise
level has a built-in lower bound, a noise floor by construction. A prior placing
noise on every mechanism would explain this, but \texttt{TabICLv2} is also
pretrained on SCMs and is less affected, so SCM pretraining alone may not
account for the effect; the way intervals are produced (ensembling,
preprocessing) may also contribute.

\paragraph{Does the prior exploit dimensional structure?}
Physical variables carry units, and by Buckingham's theorem a law of $n$
variables whose units involve $k$ independent dimensions reduces, up to a
prefactor fixed by the units, to a function of only $n-k$ dimensionless groups.
A physical problem is thus often lower-dimensional than its column count. A TFM
sees only numbers and has no direct access to this. We count the dimensionless
groups $n_\pi$ each task's inputs can form\footnote{$n_\pi = n_\mathrm{vars} -
\mathrm{rank}(D)$, with $D$ the matrix of SI dimension exponents of the input
variables, from the Feynman units table.} and ask whether a model's error
follows the raw column count or $n_\pi$. For every model it follows the column
count; once that is accounted for, $n_\pi$ adds little for most. The clearest
exception is \texttt{TabPFN-3}, whose lead over the other models is largest on
the most reducible tasks ($n_\pi{=}0$) and shrinks as $n_\pi/n_\mathrm{vars}$
grows (Fig.~\ref{fig:units}). All 26 tasks
with $n_\pi{=}0$ are pure power laws, which dimensional analysis fixes up to a
constant, so this more likely reflects their simpler form than any use of
units. We find no evidence that any model exploits dimensional structure.

\section{Conclusion} 

Tabular foundation models interpolate physical data well, without ever having
been pretrained on a physical law, and we find no detectable contamination
behind this. Their limits lie elsewhere: they become underconfident on
noiseless data and show no sign of exploiting dimensional structure --- two
properties every physical law has and no random structural causal model does.
They are excellent amortized interpolators, but a physical model would need a
prior that contains physics. Pretraining on continuous physical targets ---
absent from every corpus we audited --- would be a natural way to find out, as
would comparing them with time-series foundation models on dynamical systems.

\section*{Acknowledgments}
This work is partially supported by Schmidt Futures, a philanthropic initiative founded by Eric and Wendy Schmidt as part of the Virtual Institute for Astrophysics (VIA). The work is in part supported by computational resources provided by Calcul Qu\'ebec and the Digital Research Alliance of Canada. Y.H. and L.P. acknowledge support from the Canada Research Chairs Program, the Natural Sciences and Engineering Research Council of Canada through grants RGPIN-2020-05073 and 05102.

\bibliographystyle{plainnat}
\bibliography{refs}


\appendix

\newpage
\section{Extrapolating a damped harmonic oscillator}
\label{app:dho}

\begin{figure}[h!t]
  \centering
  \includegraphics[width=1.\hsize]{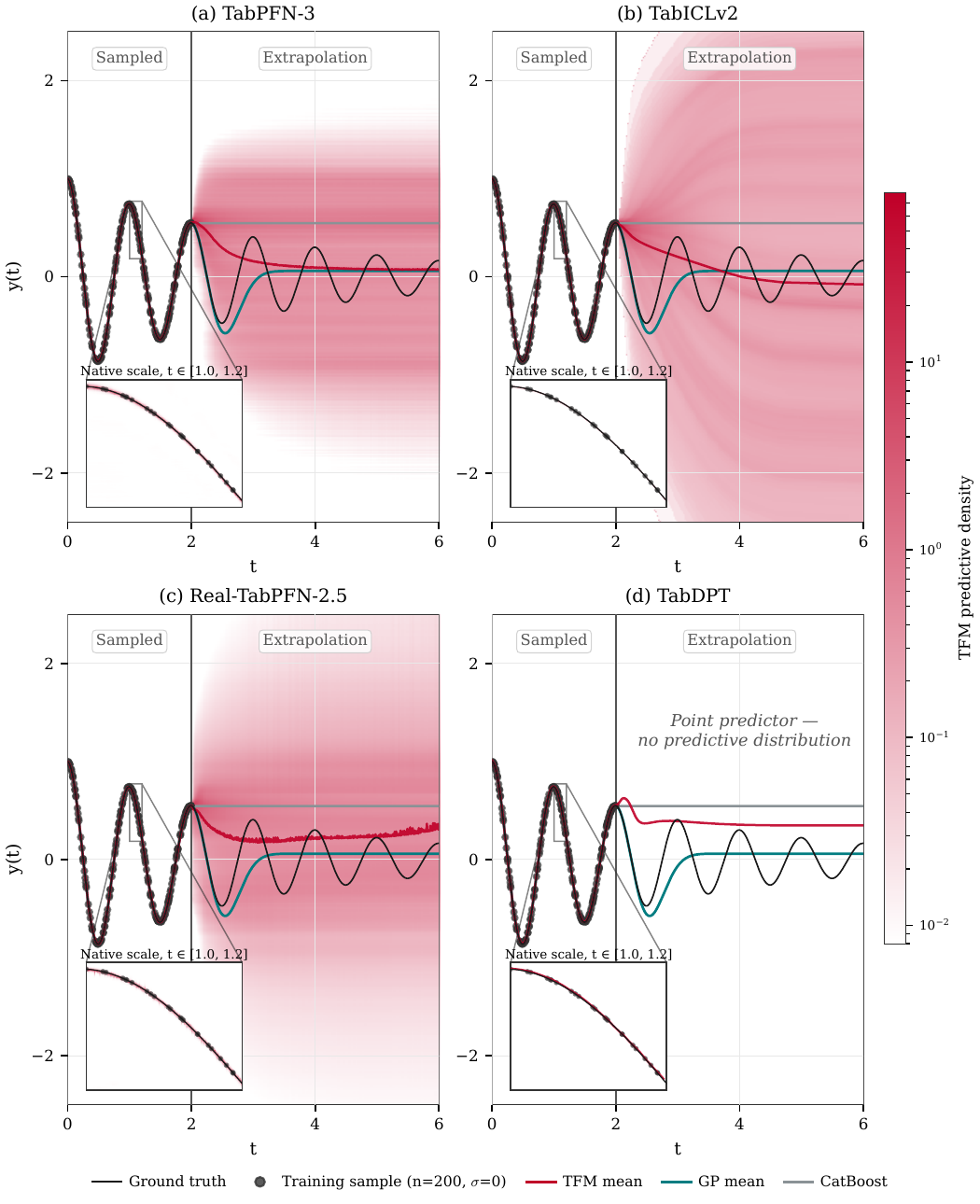}
    \caption{\textbf{Predictive distributions of all four TFMs on a damped
harmonic oscillator.} Same setup as Fig.~\ref{fig:dho}: $n=200$ noiseless
samples over $t\in[0,2]$ (points), queried to $t=6$, against a Gaussian process
(GP) and CatBoost. Color is the TFM predictive density; insets zoom an in-domain
window at native scale. Inside the densely sampled range the correct interval is close to a point: the GP's is effectively zero (${\sim}10^{-5}$, normalized), yet all three distributional TFMs report $0.010$ (\texttt{TabICLv2}) to $0.019$--$0.020$ (the two \texttt{TabPFN} variants) --- the uncertainty floor
of \S\ref{subsec:physics}, seen directly. Outside it, every model relaxes to a
constant: \texttt{TabICLv2}'s density retains an oscillatory banding,
\texttt{TabDPT} shows a partial oscillation before flattening, and
\texttt{CatBoost} holds the last observed value. \texttt{TabDPT} emits no
predictive distribution (\S\ref{sec:protocol}) and is shown as a mean only.}
  \label{fig:dho_full}
\end{figure}

We qualitatively assess how TFMs extrapolate physical dynamics by
asking them to continue a damped harmonic oscillator, a canonical physical
system, and comparing their predictions with those of a Gaussian process (GP)
and CatBoost.
Two are pretrained on synthetic SCMs only (\texttt{TabPFN-3}~\citep{TabPFN3},
\texttt{TabICLv2}~\citep{TabICL}), one on real tables only (\texttt{TabDPT}~\citep{TabDPT}), and one on both (\texttt{Real-TabPFN-2.5}~\citep{RealTabPFN}). We use $n=200$ samples and no
noise, so that any predictive width or extrapolation failure reflects the model rather than the data. Results are shown in Fig.~\ref{fig:dho_full}.

\textbf{Extrapolation.} All four models fail to continue the oscillation past
the sampled range, relaxing to a constant within roughly half a period. They do settle near the equilibrium: \texttt{TabPFN-3} and \texttt{TabICLv2} both decay toward
zero, as the true signal does, rather than holding the last observed value as
\texttt{CatBoost} does. This is consistent with~\citet{ICLRBlogPost}, who found
no evidence that \texttt{TabPFN} detects periodic structure --- and our
\texttt{TabPFN} panels are indeed a smooth, phase-agnostic blur.

\textbf{Two exceptions, and a caveat.} \texttt{TabICLv2}'s predictive density
shows wave-like banding past the boundary: the model hedges across a range of
values rather than collapsing to one, and the resulting ridges echo the scale of
the oscillation. \texttt{TabDPT} continues the signal for roughly a quarter
period before flattening. Both postdate~\citet{ICLRBlogPost} and both are
suggestive. However, a phase cross-correlation of each model's density against
the true continuation shows that the banding is \emph{not}
phase-locked\footnote{We slide the true continuation in time by 201 lags
spanning one period and measure how much of each model's predictive density
falls along each shifted curve. Had a model learned the oscillation's phase, the
unshifted curve --- the correct answer --- would score best. It does not,
ranking 98th of 201 for \texttt{TabICLv2} and 157th for \texttt{TabPFN-3}.
\texttt{TabDPT} emits no density and cannot be tested this way.}. The structure
is real multimodality, not a recovered frequency. We report it as an encouraging
direction rather than as periodic extrapolation.

\textbf{A floor even where the answer is known.} Inside the sampled range
the data are noiseless and densely sampled, so the correct predictive
distribution is close to a point mass. The GP's interval is effectively zero
there (about $10^{-5}$, normalized); unlike the default GP of \S\ref{subsec:physics}, this one
allows its noise level down to $10^{-12}$, and the optimizer drives it there.
All three distributional TFMs instead report a small but nonzero width
($0.010$--$0.020$), the same residual floor as in \S\ref{subsec:physics}.
\texttt{TabDPT} appears to escape this, but only because it emits no
predictive distribution at all (Table \ref{tab:tfms}) --- it does not report zero uncertainty,
it reports none.

\newpage
\section{Trained baselines}
\label{app:baselines}

\textbf{Trained baselines.} Table~\ref{tab:baselines} lists the six methods
fitted per task, spanning the deep, tree-based and classical families. Of
these, only the Gaussian process emits a predictive distribution; the rest
return point estimates and therefore cannot appear in any uncertainty
comparison. The GP uses an RBF kernel plus a learned white-noise term at
scikit-learn defaults, with targets normalized; its noise level is bounded below
at $10^{-5}$ of the target variance, so its 80\% interval cannot fall below about
$0.008\,\mathrm{std}(y)$ --- the noise floor discussed in \S\ref{subsec:physics}.

\begin{table}[ht]
  \centering
  \small
  \caption{\textbf{Trained baselines.} Each is fitted per task. Predictive
  output type is determined empirically from each model's implementation rather
  than from its documentation; only the Gaussian process returns a closed-form
  predictive density.}
  \begin{tabular}{lllc}
    \toprule
    Model & Family & Output & Ref. \\
    \midrule
    \texttt{RealMLP}  & MLP with tuned defaults & Point   & \citep{RealMLP} \\
    \texttt{TabM}     & MLP ensemble            & Point   & \citep{TabM} \\
    \texttt{MLP}      & Vanilla MLP             & Point   & \citep{MLP} \\
    \texttt{CatBoost} & Gradient-boosted trees  & Point   & \citep{CatBoost} \\
    \texttt{GP}       & Gaussian process        & Density & \citep{GP} \\
    \texttt{Ridge}    & Linear, $\ell_2$        & Point   & \citep{Ridge} \\
    \bottomrule
  \end{tabular}%
  \label{tab:baselines}
\end{table}

\section{Protocol details}
\label{app:protocol}

\textbf{The extrapolation shell.} Axis $j$ of a task's sampling box has range
$[l_j, h_j]$, center $c_j$ and half-width $w_j$. Test points at scale $k$ are
drawn from $[c_j - k\,w_j,\; c_j + k\,w_j]$ with rejection of any point falling
inside the $k{=}1$ box, so the test set is a shell rather than a superset of the
training domain --- without rejection, most of an OOD draw would land in-domain
and the split would measure interpolation. All $d$ axes are extended
simultaneously. We use per-axis box extension rather than a convex-hull
criterion: the latter is exponential in $d$, and these domains are boxes by
construction. The OOD region is defined from the training box alone, and it, the
noise model and the failure rule were all fixed before any result was inspected.

\textbf{Seeds in place of cross-validation.} \texttt{TabArena} and
\texttt{BeyondArena} evaluate over repeated CV folds. Because we generate data
rather than load it, the equivalent is a fresh draw per seed: each seed
independently resamples both training and test points, so no CV is needed. We
use three seeds and take the median, which is robust to a single catastrophic
draw. TFMs require multiple seeds too --- nothing is fitted, but ensemble
permutations and internal preprocessing vary between runs.

\textbf{Failure policy.} A run that runs out of memory, times out or returns
\texttt{NaN} is recorded as failed rather than dropped, and imputed at
$\mathrm{NMSE}{=}1$ in aggregates --- equivalently $R^2{=}0$, i.e.\ predicting
the training mean. This keeps a failure interpretable and model-independent
rather than letting it silently vanish from a mean. Across the full campaign, 12 of 362{,}231 runs failed permanently --- all
\texttt{TabDPT}, and all tracing to a single numerical edge case on one task at
$n_\mathrm{samples}{=}2000$, repeated across noise levels, splits and protocols.

\textbf{Validation carve-out.} Under the tuned and ensembled pipelines, the
trained baselines carve 20\% of $n_\mathrm{samples}$ for model selection and so fit
on $0.8\,n_\mathrm{samples}$; TFMs condition on all of it, since they have nothing
to select. We disclose this rather than equalize it: capping the TFM's context
to match would change the quantity being measured, and giving the baselines a
separate validation set would change the sample budget. The gap is largest at
$n_\mathrm{samples}{=}50$, where the trained models see 40 rows against the TFM's
50, and is the most likely place for a tuned-versus-default comparison to
understate the benefit of tuning.

\textbf{Preprocessing.} Column order is randomized per seed for every model.
TFMs are built to be column-permutation-invariant; checking this empirically
costs nothing and appears as a variance term. We do not disable each model's
internal preprocessing (\texttt{TabPFN} applies its own quantile and power
transforms) and record which was active in every result record: this is a
confound to disclose, not to remove. Metrics are always computed in the original
target space.

\textbf{Uncertainty evaluation.} Fig.~\ref{fig:detlim} uses the Feynman set in
domain ($k{=}1$), extended with $n_\mathrm{samples}{=}10{,}000$ on a 40-task
subset at $\sigma\in\{0, 0.1\}$. Coverage is the fraction of test points inside
the central 80\% predictive interval (10th to 90th percentiles), computed per
task, then the median is taken over seeds and over tasks.

\textbf{Cost accounting.} Wall-clock is recorded per run and split into fit and
predict phases: a TFM is nearly free to fit and costly to query, a trained model
the reverse, and only the split makes that visible. We do not attempt FLOPs,
which are not well defined across transformer, tree and kernel families and
ignore memory movement. Hardware is recorded rather than pinned, so timing
comparisons are filtered at analysis time; accuracy results are
hardware-independent. Pretraining cost is excluded by convention, as for
ImageNet-pretrained vision models.

\textbf{Reproducibility.} Each run is keyed by a hash of its resolved
configuration and skipped if the record already exists, making the campaign
idempotent and resumable. Every record embeds the full configuration, the git
commit, key package versions, hardware identifiers and per-phase timings. The
main grid comprises $316$ tasks $\times$ $3$ $n_\mathrm{samples}$ $\times$ $3$
noise levels $\times$ $2$ split scales $\times$ $3$ seeds $\times$ $10$ models
under each protocol; the tuned protocol multiplies this by 25 configurations per
trained baseline. Total measured cost was ${\approx}1300$ GPU-hours (H100 and A100) and
${\approx}4200$ CPU-hours.

\section{Per-model benchmark results}
\label{app:phyarena_full}

\begin{figure}[h!]
  \centering
  \includegraphics[width=1.\hsize]{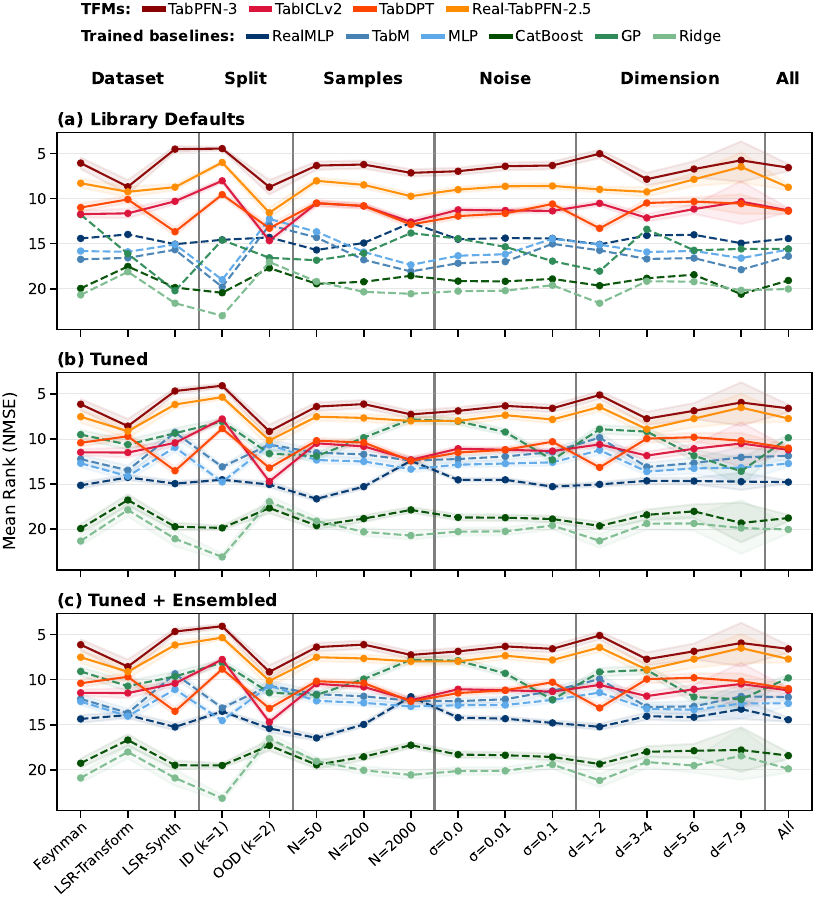}
  \caption{\textbf{Per-model results across all strata.} The same ranks as Fig.~\ref{fig:phyarena}, unpacked into all 26 model--pipeline entities and split by pipeline: \textbf{(a)} library defaults, \textbf{(b)} tuned and \textbf{(c)} tuned with post-hoc ensembling. TFMs are drawn as solid lines, trained baselines as dashed. The six trained baselines appear in every panel; the four TFMs, which have no tuning, are repeated throughout as a fixed reference, at $n_\mathrm{est}{=}1$ in (a) and $n_\mathrm{est}{=}8$ in (b) and (c). Panels share a common rank axis, so vertical gaps are comparable across them. Cells falling below 80\% task coverage are dropped rather than plotted.}
  \label{fig:phyarena_full}
\end{figure}

Figure~\ref{fig:phyarena} collapses each family to its best-performing member
under each pipeline, following \texttt{BeyondArena}'s presentation.
Figure~\ref{fig:phyarena_full} unpacks the same ranks into all 26
model--pipeline entities. This appendix records what the collapse hides, and
several observations for which there was no room in the main text.

\textbf{How the ranks are computed.} Within each (task, regime) cell we rank all
26 entities jointly --- the four TFMs at $n_\mathrm{est} \in \{1,8\}$ and the six
trained baselines at library defaults, tuned, and tuned with post-hoc ensembling
--- then average over tasks within a stratum. Ranks therefore run from 1 to 26 in
both figures, and are not comparable to a ranking computed over a smaller pool.
Cells falling below 80\% task coverage are dropped rather than plotted.

\textbf{Which model wins each family.} The collapse in Fig.~\ref{fig:phyarena}
is stable for three families of four: \texttt{TabPFN-3} is the best TFM in every
stratum under both settings, \texttt{GP} beats \texttt{Ridge} everywhere, and
\texttt{CatBoost} is the only gradient-boosted entrant. The MLP family is the
exception. \texttt{TabM} is the strongest tuned member overall, but
\texttt{RealMLP} takes over at library defaults out of domain and at the
smallest context size, and again once ensembling is applied at the largest.
Statements about "the MLP family" in the main text should be read accordingly.

\textbf{Ensembling contributes almost nothing.} Post-hoc greedy selection
improves only \texttt{RealMLP} by a measurable margin; for
\texttt{TabM}, \texttt{CatBoost}, \texttt{GP} and \texttt{Ridge} the median
change is indistinguishable from zero, with selection typically returning the
single already-best configuration. This is not a consequence of an
under-sized selection budget: allowing 25, 50, 100 or 200 rounds with
replacement over the same library yields results identical to four decimal
places, and selection saturates at four to ten distinct configurations in every
case. We attribute this to the data rather than the protocol. Ensemble selection
exploits diversity in the errors made by different configurations, and on
noiseless deterministic targets differently-tuned fits of the same model
converge on the same function and therefore make correlated errors. That
\texttt{RealMLP} is the exception is consistent with this account: its search
space varies architecture and training schedule widely enough to produce
genuinely different fits.

\textbf{Where tuning does help.} Tuning improves every trained baseline more at
the largest context size than at the smallest, as expected, and improves every
model except \texttt{Ridge} at $\sigma = 0$, where library defaults --- calibrated
for noisy real-world tables --- are furthest from appropriate.

\textbf{A note on \texttt{TabDPT}.} \texttt{TabDPT} separates from the other
TFMs specifically on \texttt{LSR-Synth} and on low-dimensional tasks, where it
performs worse than its synthetic-prior counterparts. This is not evidence of
contamination --- the effect runs in the wrong direction, since \texttt{LSR-Synth}
is the cleanest of our three sets --- but it is a real difference between models
pretrained on real corpora and those pretrained on synthetic ones, and we record
it here without an explanation.

\section{Pretraining corpora of the real-data TFMs}
\label{app:tfm_training_sets}

Both real-data-pretrained models enumerate their pretraining corpora, which is
what makes the audit in \S\ref{subsec:contamination} possible. We reproduce them
here, grouped and with live links, so that the claims in the main text can be
checked directly.

\texttt{TabDPT}'s corpus (Table~\ref{tab:corpus_tabdpt})\footnote{Reproduced from Appendix~B of \citet{TabDPT}.} comprises 122 OpenML
datasets, 93 classification and 28 regression, with domain labels assigned by
its authors. All six datasets labeled \emph{Physics/astronomy} are
classification tasks --- particle-collision and Cherenkov-shower event
discrimination, spacecraft sensor state, satellite land cover --- and none of
the nine labeled \emph{Deterministic and simulated} has a continuous target.
No regression target in the corpus is generated by an analytic function.

\texttt{Real-TabPFN-2.5}'s corpus (Table~\ref{tab:corpus_realtabpfn})\footnote{Reproduced from Appendix~C.1 of \citet{RealTabPFN25}.} comprises
43 datasets from OpenML and Kaggle. It contains no real-world regression table
at all: its one function-generated entry, \texttt{fried}, links to OpenML~901,
the binarized variant of Friedman~\#1 in which the continuous target has been
replaced by a two-class label thresholded at its mean. Two Sloan Digital Sky
Survey catalogs appear, both as star/galaxy/quasar classification.

\begin{table}[h!]
\centering
\tiny
\setlength{\tabcolsep}{4pt}
\caption{\textbf{\texttt{TabDPT} pretraining corpus} (122 datasets), grouped by
the domain labels assigned in its own appendix and sorted alphabetically within
group. Regression targets are marked $^{\dagger}$; all others are
classification, except \texttt{Census-Income}$^{*}$, for which no target type is
given. No regression target in the corpus is generated by an analytic function,
and every dataset in the \emph{Physics/astronomy} and \emph{Deterministic and
simulated} groups is a classification task.}
\label{tab:corpus_tabdpt}
\begin{tabularx}{\hsize}{XXX}
\toprule
\multicolumn{3}{l}{\emph{Physics/astronomy} --- 6 datasets, 0 regression} \\[1pt]
\href{https://www.openml.org/d/23512}{\texttt{higgs}} & \href{https://www.openml.org/d/40679}{\texttt{magic}} & \href{https://www.openml.org/d/1120}{\texttt{MagicTelescope}} \\
\href{https://www.openml.org/d/41150}{\texttt{MiniBooNE}} & \href{https://www.openml.org/d/40900}{\texttt{Satellite}} & \href{https://www.openml.org/d/40685}{\texttt{shuttle}} \\
\addlinespace[3pt]
\multicolumn{3}{l}{\emph{Deterministic and simulated} --- 9 datasets, 0 regression} \\[1pt]
\href{https://www.openml.org/d/1459}{\texttt{artificial-characters}} & \href{https://www.openml.org/d/1118}{\texttt{chess}} & \href{https://www.openml.org/d/1479}{\texttt{hill-valley}} \\
\href{https://www.openml.org/d/1481}{\texttt{kr-vs-k}} & \href{https://www.openml.org/d/184}{\texttt{kropt}} & \href{https://www.openml.org/d/40706}{\texttt{parity5\_plus\_5}} \\
\href{https://www.openml.org/d/1567}{\texttt{poker-hand}} & \href{https://www.openml.org/d/816}{\texttt{puma8NH}} & \href{https://www.openml.org/d/1507}{\texttt{twonorm}} \\
\addlinespace[3pt]
\multicolumn{3}{l}{\emph{Other science} --- 4 datasets, 1 regression} \\[1pt]
\href{https://www.openml.org/d/1476}{\texttt{gas-drift}} & \href{https://www.openml.org/d/1477}{\texttt{gas-drift-different-concentrations}} & \href{https://www.openml.org/d/1116}{\texttt{musk}} \\
\href{https://www.openml.org/d/44145}{\texttt{sulfur}}$^{\dagger}$ &  &  \\
\addlinespace[3pt]
\multicolumn{3}{l}{\emph{Biology/ecology} --- 11 datasets, 1 regression} \\[1pt]
\href{https://www.openml.org/d/23380}{\texttt{cjs}} & \href{https://www.openml.org/d/1596}{\texttt{covertype}} & \href{https://www.openml.org/d/40646}{\texttt{GAMETES\_Epistasis\_2-Way\_20atts}} \\
\href{https://www.openml.org/d/24}{\texttt{mushroom}} & \href{https://www.openml.org/d/1493}{\texttt{one-hundred-plants-texture}} & \href{https://www.openml.org/d/1142}{\texttt{OVA\_Endometrium}} \\
\href{https://www.openml.org/d/1130}{\texttt{OVA\_Lung}} & \href{https://www.openml.org/d/871}{\texttt{pollen}} & \href{https://www.openml.org/d/1036}{\texttt{sylva\_agnostic}} \\
\href{https://www.openml.org/d/44056}{\texttt{visualizing\_soil}}$^{\dagger}$ & \href{https://www.openml.org/d/40733}{\texttt{yeast}} &  \\
\addlinespace[3pt]
\multicolumn{3}{l}{\emph{Medical/human sensor} --- 11 datasets, 4 regression} \\[1pt]
\href{https://www.openml.org/d/1466}{\texttt{cardiotocography}} & \href{https://www.openml.org/d/45022}{\texttt{Diabetes130US}} & \href{https://www.openml.org/d/1471}{\texttt{eeg-eye-state}} \\
\href{https://www.openml.org/d/1044}{\texttt{eye\_movements}} & \href{https://www.openml.org/d/1483}{\texttt{ldpa}} & \href{https://www.openml.org/d/802}{\texttt{pbcseq}} \\
\href{https://www.openml.org/d/3277}{\texttt{QSAR-TID-10980}}$^{\dagger}$ & \href{https://www.openml.org/d/3050}{\texttt{QSAR-TID-11}}$^{\dagger}$ & \href{https://www.openml.org/d/45041}{\texttt{topo\_2\_1}}$^{\dagger}$ \\
\href{https://www.openml.org/d/1509}{\texttt{walking-activity}} & \href{https://www.openml.org/d/45032}{\texttt{yprop\_4\_1}}$^{\dagger}$ &  \\
\addlinespace[3pt]
\multicolumn{3}{l}{\emph{Financial/demographic} --- 15 datasets, 2 regression} \\[1pt]
\href{https://www.openml.org/d/1043}{\texttt{ada\_agnostic}} & \href{https://www.openml.org/d/1037}{\texttt{ada\_prior}} & \href{https://www.openml.org/d/4535}{\texttt{Census-Income}}$^{*}$ \\
\href{https://www.openml.org/d/44089}{\texttt{credit}} & \href{https://www.openml.org/d/45020}{\texttt{default-of-credit-card-clients}} & \href{https://www.openml.org/d/45026}{\texttt{heloc}} \\
\href{https://www.openml.org/d/821}{\texttt{house\_16H}} & \href{https://www.openml.org/d/843}{\texttt{house\_8L}} & \href{https://www.openml.org/d/42563}{\texttt{house\_prices\_nominal}}$^{\dagger}$ \\
\href{https://www.openml.org/d/382}{\texttt{ipums\_la\_97-small}} & \href{https://www.openml.org/d/1002}{\texttt{ipums\_la\_98-small}} & \href{https://www.openml.org/d/1018}{\texttt{ipums\_la\_99-small}} \\
\href{https://www.openml.org/d/981}{\texttt{kdd\_internet\_usage}} & \href{https://www.openml.org/d/1568}{\texttt{nursery}} & \href{https://www.openml.org/d/42730}{\texttt{us\_crime}}$^{\dagger}$ \\
\addlinespace[3pt]
\multicolumn{3}{l}{\emph{Human behaviour} --- 17 datasets, 8 regression} \\[1pt]
\href{https://www.openml.org/d/44063}{\texttt{Bike\_Sharing\_Demand}}$^{\dagger}$ & \href{https://www.openml.org/d/41540}{\texttt{black\_friday}}$^{\dagger}$ & \href{https://www.openml.org/d/4549}{\texttt{Buzzinsocialmedia\_Twitter}}$^{\dagger}$ \\
\href{https://www.openml.org/d/41434}{\texttt{Click\_prediction\_small}} & \href{https://www.openml.org/d/45039}{\texttt{compas-two-years}} & \href{https://www.openml.org/d/43072}{\texttt{KDDCup09-Upselling}} \\
\href{https://www.openml.org/d/1111}{\texttt{KDDCup09\_appetency}} & \href{https://www.openml.org/d/44065}{\texttt{nyc-taxi-green-dec-2016}}$^{\dagger}$ & \href{https://www.openml.org/d/42734}{\texttt{okcupid-stem}} \\
\href{https://www.openml.org/d/42724}{\texttt{OnlineNewsPopularity}}$^{\dagger}$ & \href{https://www.openml.org/d/42742}{\texttt{porto-seguro}} & \href{https://www.openml.org/d/45038}{\texttt{road-safety}} \\
\href{https://www.openml.org/d/42572}{\texttt{Santander\_transaction\_value}}$^{\dagger}$ & \href{https://www.openml.org/d/45043}{\texttt{seattlecrime6}}$^{\dagger}$ & \href{https://www.openml.org/d/42732}{\texttt{sf-police-incidents}} \\
\href{https://www.openml.org/d/40536}{\texttt{SpeedDating}} & \href{https://www.openml.org/d/44136}{\texttt{wine\_quality}}$^{\dagger}$ &  \\
\addlinespace[3pt]
\multicolumn{3}{l}{\emph{Industrial/operational} --- 10 datasets, 4 regression} \\[1pt]
\href{https://www.openml.org/d/1169}{\texttt{airlines}} & \href{https://www.openml.org/d/42728}{\texttt{Airlines\_DepDelay\_10M}}$^{\dagger}$ & \href{https://www.openml.org/d/45046}{\texttt{Allstate\_Claims\_Severity}}$^{\dagger}$ \\
\href{https://www.openml.org/d/4135}{\texttt{Amazon\_employee\_access}} & \href{https://www.openml.org/d/41138}{\texttt{APSFailure}} & \href{https://www.openml.org/d/45045}{\texttt{delays\_zurich\_transport}}$^{\dagger}$ \\
\href{https://www.openml.org/d/1112}{\texttt{KDDCup09\_churn}} & \href{https://www.openml.org/d/41162}{\texttt{kick}} & \href{https://www.openml.org/d/44061}{\texttt{Mercedes\_Benz\_Greener\_Manufacturing}}$^{\dagger}$ \\
\href{https://www.openml.org/d/44122}{\texttt{pol}} &  &  \\
\addlinespace[3pt]
\multicolumn{3}{l}{\emph{Computing} --- 4 datasets, 3 regression} \\[1pt]
\href{https://www.openml.org/d/42746}{\texttt{KDDCup99}} & \href{https://www.openml.org/d/43071}{\texttt{MIP-2016-Reg.}}$^{\dagger}$ & \href{https://www.openml.org/d/41980}{\texttt{SAT11-HAND-runtime-Reg.}}$^{\dagger}$ \\
\href{https://www.openml.org/d/44069}{\texttt{SGEMM\_GPU\_kernel\_performance}}$^{\dagger}$ &  &  \\
\addlinespace[3pt]
\multicolumn{3}{l}{\emph{Vision/audio/text features} --- 7 datasets, 0 regression} \\[1pt]
\href{https://www.openml.org/d/1457}{\texttt{amazon-commerce-reviews}} & \href{https://www.openml.org/d/389}{\texttt{fbis.wc}} & \href{https://www.openml.org/d/375}{\texttt{JapaneseVowels}} \\
\href{https://www.openml.org/d/396}{\texttt{la1s.wc}} & \href{https://www.openml.org/d/30}{\texttt{page-blocks}} & \href{https://www.openml.org/d/312}{\texttt{scene}} \\
\href{https://www.openml.org/d/1503}{\texttt{spoken-arabic-digit}} &  &  \\
\addlinespace[3pt]
\multicolumn{3}{l}{\emph{Other or not provided} --- 28 datasets, 5 regression} \\[1pt]
\href{https://www.openml.org/d/41156}{\texttt{ada}} & \href{https://www.openml.org/d/44137}{\texttt{Ailerons}}$^{\dagger}$ & \href{https://www.openml.org/d/41147}{\texttt{albert}} \\
\href{https://www.openml.org/d/966}{\texttt{analcatdata\_halloffame}} & \href{https://www.openml.org/d/44055}{\texttt{analcatdata\_supreme}}$^{\dagger}$ & \href{https://www.openml.org/d/41142}{\texttt{christine}} \\
\href{https://www.openml.org/d/42727}{\texttt{colleges}}$^{\dagger}$ & \href{https://www.openml.org/d/930}{\texttt{colleges\_usnews}} & \href{https://www.openml.org/d/41163}{\texttt{dilbert}} \\
\href{https://www.openml.org/d/41167}{\texttt{dionis}} & \href{https://www.openml.org/d/846}{\texttt{elevators}} & \href{https://www.openml.org/d/41164}{\texttt{fabert}} \\
\href{https://www.openml.org/d/41159}{\texttt{guillermo}} & \href{https://www.openml.org/d/41169}{\texttt{helena}} & \href{https://www.openml.org/d/273}{\texttt{IMDB.drama}} \\
\href{https://www.openml.org/d/41168}{\texttt{jannis}} & \href{https://www.openml.org/d/41143}{\texttt{jasmine}} & \href{https://www.openml.org/d/41144}{\texttt{madeline}} \\
\href{https://www.openml.org/d/40680}{\texttt{mofn-3-7-10}} & \href{https://www.openml.org/d/44068}{\texttt{particulate-matter-ukair-2017}}$^{\dagger}$ & \href{https://www.openml.org/d/41145}{\texttt{philippine}} \\
\href{https://www.openml.org/d/1453}{\texttt{PieChart3}} & \href{https://www.openml.org/d/1444}{\texttt{PizzaCutter3}} & \href{https://www.openml.org/d/41161}{\texttt{riccardo}} \\
\href{https://www.openml.org/d/41165}{\texttt{robert}} & \href{https://www.openml.org/d/41146}{\texttt{sylvine}} & \href{https://www.openml.org/d/41166}{\texttt{volkert}} \\
\href{https://www.openml.org/d/42705}{\texttt{Yolanda}}$^{\dagger}$ &  &  \\
\addlinespace[3pt]\bottomrule
\end{tabularx}
\end{table}

\begin{table}[h!]
\centering
\tiny
\setlength{\tabcolsep}{4pt}
\caption{\textbf{\texttt{Real-TabPFN-2.5} pretraining corpus} (43 datasets), as
listed in its technical report, sorted alphabetically within source. Every entry
is a classification task: the corpus contains no real-world regression table.
Its one function-generated entry, \texttt{fried}, links to OpenML~901 --- the
binarised variant of Friedman~\#1, in which the continuous target has been
replaced by a two-class label thresholded at its mean.}
\label{tab:corpus_realtabpfn}
\begin{tabularx}{\hsize}{XXX}
\toprule
\multicolumn{3}{l}{\emph{OpenML} --- 20 datasets} \\[1pt]
\href{https://www.openml.org/d/1459}{\texttt{artificial-characters}} & \href{https://www.openml.org/d/251}{\texttt{BNG(breast-w)}} & \href{https://www.openml.org/d/137}{\texttt{BNG(tic-tac-toe)}} \\
\href{https://www.openml.org/d/40668}{\texttt{connect\_4}} & \href{https://www.openml.org/d/1471}{\texttt{eeg-eye-state}} & \href{https://www.openml.org/d/43551}{\texttt{Employee-Turnover-at-TECHCO}} \\
\href{https://www.openml.org/d/1044}{\texttt{eye\_movements}} & \href{https://www.openml.org/d/41787}{\texttt{FOREX\_eurpln-hour-High}} & \href{https://www.openml.org/d/901}{\texttt{fried}} \\
\href{https://www.openml.org/d/1476}{\texttt{gas-drift}} & \href{https://www.openml.org/d/23512}{\texttt{higgs}} & \href{https://www.openml.org/d/43039}{\texttt{Internet Firewall Data}} \\
\href{https://www.openml.org/d/44201}{\texttt{Intersectional-Bias-Assessment-(Training-Data)}} & \href{https://www.openml.org/d/43904}{\texttt{law-school-admission-binary}} & \href{https://www.openml.org/d/43617}{\texttt{Medical-Appointment}} \\
\href{https://www.openml.org/d/41671}{\texttt{microaggregation2}} & \href{https://www.openml.org/d/43923}{\texttt{mushroom}} & \href{https://www.openml.org/d/44226}{\texttt{NewspaperChurn}} \\
\href{https://www.openml.org/d/1568}{\texttt{nursery}} & \href{https://www.openml.org/d/46676}{\texttt{WBCAtt}} &  \\
\addlinespace[3pt]
\multicolumn{3}{l}{\emph{Kaggle} --- 23 datasets} \\[1pt]
\href{https://www.kaggle.com/datasets/himselfthedecker/aam-avaliacao-dataset}{\texttt{aam\_avaliacao\_dataset}} & \href{https://www.kaggle.com/datasets/rohanshetty678/air-traffic-data}{\texttt{Air Traffic Data}} & \href{https://www.kaggle.com/datasets/stefadp/ansibledefectsprediction}{\texttt{ansible-defects-prediction}} \\
\href{https://www.kaggle.com/datasets/nehaprabhavalkar/av-healthcare-analytics-ii}{\texttt{AV Healthcare Analytics II}} & \href{https://www.kaggle.com/datasets/tarunchilkur/client}{\texttt{Candidate Selection}} & \href{https://www.kaggle.com/datasets/sulianova/cardiovascular-disease-dataset}{\texttt{Cardio Disease}} \\
\href{https://www.kaggle.com/datasets/aniketng21600/crop-damage-information-in-india}{\texttt{Classification - Crop Damages in India (2015-2019)}} & \href{https://www.kaggle.com/datasets/christianlillelund/csgo-round-winner-classification}{\texttt{CSGO Round Winner Classification}} & \href{https://www.kaggle.com/datasets/vpkprasanna/flower-type-prediction-machine-hack}{\texttt{Flower Type Prediction Machine Hack}} \\
\href{https://www.kaggle.com/datasets/gunner38/horseracing/data}{\texttt{Horse Racing - Tipster Bets}} & \href{https://www.kaggle.com/datasets/kanuriviveknag/road-accidents-severity-dataset}{\texttt{How severe the accident could be}} & \href{https://www.kaggle.com/datasets/pankeshpatel/hrcommasep}{\texttt{hr-comma-sep}} \\
\href{https://www.kaggle.com/datasets/jsrojas/ip-network-traffic-flows-labeled-with-87-apps}{\texttt{ip-network-traffic-flows-labeled-with-87-apps}} & \href{https://www.kaggle.com/datasets/pawan2905/jantahack-cross-sell-prediction}{\texttt{Janatahack cross-sell prediction}} & \href{https://www.kaggle.com/datasets/mamtadhaker/lt-vehicle-loan-default-prediction}{\texttt{L\&{}T Vehicle Loan Default Prediction}} \\
\href{https://www.kaggle.com/datasets/benfattori/league-of-legends-diamond-games-first-15-minutes}{\texttt{League of Legends Diamond Games (First 15 Minutes)}} & \href{https://www.kaggle.com/code/franciscoescobar/richter-s-predictor-modeling-earthquake-damage}{\texttt{Richter's Predictor Modeling Earthquake Damage}} & \href{https://www.kaggle.com/datasets/kartikjaspal/server-logs-suspicious}{\texttt{Server Logs - Suspicious}} \\
\href{https://www.kaggle.com/datasets/lucidlenn/sloan-digital-sky-survey}{\texttt{Sloan Digital Sky Survey DR14}} & \href{https://www.kaggle.com/datasets/muhakabartay/sloan-digital-sky-survey-dr16}{\texttt{Sloan Digital Sky Survey DR16}} & \href{https://www.kaggle.com/datasets/brajeshmohapatra/term-deposit-prediction-data-set}{\texttt{Term Deposit Prediction Data Set}} \\
\href{https://www.kaggle.com/datasets/danielamigo/trajectorybasedshipclassification/data}{\texttt{trajectory-based-ship-classification}} & \href{https://www.kaggle.com/datasets/mhdzahier/travel-insurance}{\texttt{Travel Insurance}} &  \\
\addlinespace[3pt]\bottomrule
\end{tabularx}
\end{table}


\end{document}